\documentclass[letterpaper, 10 pt, conference]{ieeeconf}
\IEEEoverridecommandlockouts
\usepackage{amsmath,amssymb}

\usepackage{amsthm}
\usepackage{graphicx}
\usepackage{booktabs}
\usepackage{cite}
\usepackage{url}
\usepackage{algorithm}
\usepackage{algpseudocode}
\usepackage{tikz}
\usepackage[hidelinks]{hyperref}
\usetikzlibrary{calc}

\theoremstyle{definition}
\newtheorem{definition}{Definition}
\newtheorem{theorem}[definition]{Theorem}
\newtheorem{lemma}[definition]{Lemma}
\newtheorem{proposition}[definition]{Proposition}
\newtheorem{corollary}[definition]{Corollary}

\newcommand{\R}{\mathbb{R}}
\newcommand{\X}{\mathcal{X}}
\newcommand{\Zc}{\mathcal{Z}}
\newcommand{\Zv}{\mathcal{Z}_v}
\newcommand{\D}{\mathcal{D}}
\newcommand{\Ob}{\mathcal{O}}
\newcommand{\er}{\mathbf{e}_r}
\newcommand{\evert}{\mathbf{e}_3}
\newcommand{\bb}{\mathbf{b}}
\newcommand{\ww}{\mathbf{w}}
\newcommand{\vv}{\mathbf{v}}

\newcommand{\Lf}{L_f}
\newcommand{\Lg}{L_g}
\newcommand{\vperp}{v_\perp^2}

\title{\LARGE \bf
Transferable Neural CBFs via Generalized Closure}

\author{Nishant Jayesh Bhave, Parthib Khound%
\thanks{N. J. Bhave and P. Khound are with the Indian Institute of Technology Bombay, Mumbai 400076, India
{\tt\small {nishant.bhave, khound}@iitb.ac.in}}%
}

\begin{document}

% IEEE preprint notice in the top margin of page 1, added at shipout so it moves no text.
\AddToHookNext{shipout/foreground}{\put(0,-0.3in){\makebox[\paperwidth][c]{\parbox[t]{6.8in}{\centering\footnotesize This work has been submitted to the IEEE for possible publication.
Copyright may be transferred without notice, after which this version may no longer be
accessible.}}}}
\maketitle
\enlargethispage{24pt}
\thispagestyle{empty}
\pagestyle{empty}

\begin{abstract}
Classical control barrier functions (CBFs) provide safety constraints for controllers, but designing
them by hand is difficult for underactuated vehicles and the resulting safe sets can be overly
conservative. Neural CBFs can learn more informative barriers from data, yet a learned barrier is
usually verified only for one fixed obstacle configuration. We study whether one neural barrier can
be reused across environments by expressing the state relative to each obstacle. The key gap is that
these relative coordinates may discard information needed to predict their own evolution. We give a
generalized closure condition that represents this missing information as a residual and a lifting
theorem that transfers a verified relative-coordinate barrier to every obstacle in a family. A rank
test also identifies when symmetry alone cannot justify the transfer. We validate the conditions on
a double integrator, a planar aircraft, and a 12-state quadrotor. Coordinates that fail the test
lead to obstacle penetration or to a barrier condition that fails near hover, whereas body-frame
coordinates close exactly for the quadrotor. The resulting barrier transfers to new maps without
retraining and is effective in the reported deployment and fast-start experiments.
\end{abstract}

%%%%%%%%%%%%%%%%%%%%%%%%%%%%%%%%%%%%%%%%%%%%%%%%%%%%%%%%%%%%%%%%%%%%%%%%%%%%%%%%
\section{INTRODUCTION}

Control barrier functions (CBFs) turn a safety specification into a constraint on the input that a
quadratic program enforces online \cite{ames2017cbfqp}, and the same program can filter the
output of any nominal controller \cite{gurriet2018backup}. The barrier
itself must be constructed, and that is the bottleneck. For systems with bounded inputs in which the
input reaches the obstacle constraint only after several integrations, high-order CBFs
\cite{xiao2019hocbf} give a systematic construction, but the designer chooses the function and its
gains, and nothing guarantees that the resulting constraint can be met within the input bounds. On
underactuated vehicles, whose ability to stop depends on attitude and rates as well as on distance
and velocity, such a barrier is hard to make both valid and informative. Hamilton--Jacobi reachability computes the
largest safe set, the viability kernel \cite{aubin1991viability}, on a grid, at a cost
exponential in the state dimension \cite{bansal2017hj}, and learned value functions extend it to higher dimension for a fixed environment and without
formal guarantees \cite{bansal2021deepreach}.

A third approach is to learn the barrier from its defining inequalities \cite{dawson2023survey}. A
learned function is a certificate only if those inequalities hold, and a growing line of work verifies
learned barriers formally
\cite{zhang2023exact,hu2024symbolic,vertovec2025scalable,tayal2024verified}. In each of these works, the
verified object is one barrier for one fixed environment, verified in full-state coordinates, so that
any change of the environment requires retraining and reverification.

Shared and compositional constructions remove the dependence on the environment. A shared network is
evaluated on the local configuration of each agent \cite{qin2021decentralized,zhang2025gcbfplus}, or
one network $h_1$ is trained in coordinates relative to a single obstacle, evaluated once for each
obstacle of a map, and combined by a smooth minimum \cite{derajic2026cncbf} or, as here, by one
constraint per obstacle. The object to verify then lives on
the relative coordinates of one obstacle, whatever the number and placement of the obstacles. Such
networks transfer across maps in practice, but none of these constructions is formally verified, so
the transfer rests on conditions that are never checked for the learned function. Equivariance results evaluate the barrier at a transformed full state, so no information is
lost \cite{wiltz2025equivariances}, but coordinates chosen for learning often discard information on
which the dynamics depend.

This paper states when the per-obstacle construction is valid. For the barrier condition to transfer,
the relative coordinates must
determine their own time derivative, up to a residual whose range they determine, and then a barrier
condition on $h_1$ that holds on a compact set of relative coordinates holds for every obstacle in
every map. A rank test shows in advance when symmetry alone cannot guarantee this. Coordinates that
fail it penetrate obstacles on a planar aircraft and fail the condition near hover on a 12-state
quadrotor. Body-frame coordinates pass it, and the barrier learned in them is deployed on new maps
without retraining. We check the hypotheses on sampled boxes and states, so the result is a
statistical estimate of where they hold, not a proof over the whole domain, and the deployed
controller adds discrete-time safeguards that the theory does not cover. The contributions are the
following:
\enlargethispage{12pt}
\begin{enumerate}
\setlength{\topsep}{-6pt}
\item Generalized closure and a lifting theorem: a per-obstacle barrier condition that holds on a
compact set of relative coordinates holds for every obstacle in every map, and if one input meets all
obstacle constraints the state stays safe until it leaves the lifted domain, with the missing
information made explicit as a residual (Theorem~\ref{thm:lift},
Corollary~\ref{cor:safety}).
\item A rank condition, comparing the rank of the coordinates with $N=n+p-\dim G$, that is necessary for symmetry to guarantee closure and can be checked before training, and explicit residual sets for three systems
(Proposition~\ref{prop:count}, Table~\ref{tab:count}).
\item Body-frame relative coordinates that close exactly for the full 12-state quadrotor, and a barrier
learned in them, checked at sampled states, and deployed without retraining on maps not seen in
training
(Proposition~\ref{prop:body}, Section~\ref{sec:results}).
\end{enumerate}

%%%%%%%%%%%%%%%%%%%%%%%%%%%%%%%%%%%%%%%%%%%%%%%%%%%%%%%%%%%%%%%%%%%%%%%%%%%%%%%%
\section{RELATED WORK}\label{sec:related}

\paragraph*{Learned barriers and their verification}
Neural certificates are trained by penalizing violations of their defining inequalities at sampled
states \cite{dawson2023survey}, which alone does not make them certificates. Zhang et al.\ verify ReLU barriers exactly by exploiting the
piecewise-linear structure of the network \cite{zhang2023exact}. Hu et al.\ propagate symbolic bounds
through the gradient of ReLU barriers and their Lie derivatives \cite{hu2024symbolic}, and Vertovec et
al.\ extend linear bound propagation to the gradient of smooth networks \cite{vertovec2025scalable},
both in the family of certified bound-propagation (CROWN) relaxations
\cite{zhang2018crown} on which our verifier is also built.
Tayal et al.\ train barriers for stochastic systems that are valid by construction through Lipschitz
bounds on the network and its derivatives \cite{tayal2024verified}, and give probabilistic guarantees
for deterministic systems through conformal prediction \cite{tayal2025cpncbf}. We check in relative coordinates instead, statistically and once for an obstacle family, and
lift the result to every map by a theorem.

\paragraph*{Shared and compositional barriers}
Qin et al.\ learn decentralized barrier certificates whose network takes an agent's own state and
the states of its neighbours within an observation radius, with the trained parameters reused for
every agent \cite{qin2021decentralized}. GCBF+ generalizes this with graph neural
networks over neighbours and sensed obstacles \cite{zhang2025gcbfplus}. Closest to the present work,
CN-CBF learns one network relative to a single, possibly moving, obstacle from reachability data,
evaluates it for every obstacle, and composes the results with a smooth minimum
\cite{derajic2026cncbf}. It writes the relative state of a unicycle in the robot frame, observing that
the world-frame relative dynamics are not a function of the relative state. We make that observation
a condition, give a test for it, and handle the case in which it fails. In these works the learned certificate
is not verified, so transfer is observed in experiments, although GCBF+ proves that a function meeting
its graph CBF conditions certifies safety for any number of agents. CN-CBF computes its training targets by reachability on a four-dimensional relative grid, which
would be twelve-dimensional for the full quadrotor here. The minimum of barriers is a nonsmooth
barrier whenever one input meets every active constraint \cite{glotfelter2017nonsmooth}, and whether
such an input exists within the input bounds is a separate question \cite{breeden2023compositions},
which we assume rather than solve.

\paragraph*{Symmetry and reduction}
Control systems with symmetries reduce to their quotient \cite{grizzle1985symmetries}. Wiltz and
Dimarogonas show that when the dynamics are equivariant, a CBF computed for one constraint yields CBFs
for its transformations, and that when the constraint is also symmetric a reachability-based CBF need
only be computed on a subset \cite{wiltz2025equivariances}.  We address coordinates that do not determine the state
up to symmetry, as coordinates chosen for learning often do not, make the
missing information an explicit residual whose range the coordinates determine, and give a rank test
that reveals the shortfall in advance.

\paragraph*{Backup sets}
Backup control barrier functions replace the viability kernel by the states from
which a fixed backup policy keeps the system safe, an inner approximation that is checked online by
forward simulation and enforced along the predicted trajectory inside the QP \cite{chen2021backup},
and policy neural
CBFs learn the barrier as the value of a fixed policy, without verification \cite{so2024pncbf}. We use
such rollouts in two roles, to label the training data and as a discrete-time safeguard in
deployment.

%%%%%%%%%%%%%%%%%%%%%%%%%%%%%%%%%%%%%%%%%%%%%%%%%%%%%%%%%%%%%%%%%%%%%%%%%%%%%%%%
\section{PRELIMINARIES}\label{sec:prelim}

We consider $\dot x=f(x)+g(x)u$ with $x\in\X\subseteq\R^n$ open, $f,g$ continuously
differentiable, and $u$ in a compact box $U=[u_{\min},u_{\max}]\subset\R^m$. For a continuously
differentiable $h$ write $\Lf h=\nabla h^\top f$ and $\Lg h=\nabla h^\top g$.

\begin{definition}[CBF on a band]\label{def:zcbf}
For $\gamma,\delta,\eta>0$, $h$ is a CBF on $\{-\delta<h<\eta\}$ if
\begin{equation}\label{eq:cbf}
\max_{u\in U}\big[\Lf h(x)+\Lg h(x)\,u\big]\ \ge\ -\gamma\,h(x)
\end{equation}
whenever $-\delta<h(x)<\eta$.
\end{definition}
This is the zeroing CBF condition of \cite{ames2017cbfqp} with a linear class-$\mathcal K$ function,
imposed only on a band around the boundary. The band is an open neighbourhood of $\{h=0\}$, which is
all that forward invariance of $\{h\ge0\}$ needs (Corollary~\ref{cor:safety}).

Since the bracket is affine in $u$, $\max_{u\in U}au=\sum_j\max(a_ju_{\min,j},a_ju_{\max,j})$ makes
\eqref{eq:cbf} explicit in $\Lf h$ and $\Lg h$.

Obstacles are spheres $\theta=(c,r)\in\Theta\subseteq\R^p$, where $c\in\R^d$ is the centre,
$r>0$ is the radius, and $p=d+1$. With $P$ selecting position and a margin
$m_g>0$, the inflated obstacle is $\Ob_\theta=\{x:\lVert Px-c\rVert<r+m_g\}$, and a map is a finite set
$\{\theta_i\}_{i=1}^K$. We seek one $h_1$ and coordinates $z$ such that, for every $\theta$,
$h_\theta=h_1\circ z_\theta$ is a CBF excluding $\Ob_\theta$, established by one check that
depends on neither $\theta$ nor the map.

%%%%%%%%%%%%%%%%%%%%%%%%%%%%%%%%%%%%%%%%%%%%%%%%%%%%%%%%%%%%%%%%%%%%%%%%%%%%%%%%
\section{CERTIFICATES UNDER CLOSURE}\label{sec:closure}

Let $\D=\{(x,\theta):Px\neq c\}$. Relative coordinates are a continuously differentiable map
$z:\D\to\R^q$, $z_\theta(x)=z(x,\theta)$, whose first component is the signed distance
$s=\lVert Px-c\rVert-r$, so that $x\in\Ob_\theta$ if and only if $s<m_g$. Let $\Zc=z(\D)$.

\begin{definition}[Closure]\label{def:closure}
The coordinates close exactly if there is a continuous $F:\Zc\times U\to\R^q$ with
$D_xz_\theta(x)(f(x)+g(x)u)=F(z_\theta(x),u)$ for all $(x,\theta)\in\D$ and $u\in U$.
\end{definition}

\begin{definition}[Generalized closure]\label{def:genclosure}
The coordinates close up to a residual if there are a set-valued map $\Xi:\Zc\rightrightarrows\R^\ell$
with nonempty compact values and a continuous $F:\Zc\times U\times\R^\ell\to\R^q$, affine in $u$,
such that for every $(x,\theta)\in\D$ some $\xi^\star\in\Xi(z_\theta(x))$ satisfies
\begin{equation}\label{eq:genclosure}
D_xz_\theta(x)\big(f(x)+g(x)u\big)=F\big(z_\theta(x),u,\xi^\star\big)\quad\text{for all }u\in U.
\end{equation}
\end{definition}

The residual $\xi^\star$ collects what $\dot z$ depends on but $z$ does not determine, such as the
sign of the tangential velocity for the planar aircraft of Section~\ref{sec:pvtol}. It is fixed by
the state, not by the input, so the input may depend on it, and the certificate condition below takes the minimum over the
residual outside the maximum over the input. Exact closure is the case of singleton $\Xi$. On a bounded set of states any system satisfies Definition~\ref{def:genclosure} trivially, with the full state as residual, and the definition is
useful only when $\Xi(z)$ is explicit and computable from $z$.

\begin{definition}[Relative certificate]\label{def:cert}
Let $\Zv\subseteq\Zc$ be compact. A continuously differentiable $h_1$ is a relative certificate on $\Zv$ if (V1) $h_1(z)<0$ for $z\in\Zv$ with $s(z)<m_g$, and (V2) for $z\in\Zv$ with
$-\delta<h_1(z)<\eta$,
\begin{equation}\label{eq:v2}
\Psi(z):=\min_{\xi\in\Xi(z)}\max_{u\in U}\nabla h_1(z)^\top F(z,u,\xi)+\gamma h_1(z)\ \ge\ 0 .
\end{equation}
\end{definition}
In this paper (V1) and (V2) are checked statistically, on sampled boxes or states, so a trained
network is shown to meet Definition~\ref{def:cert} only on the part of $\Zv$ that the checks cover.

\begin{theorem}[Lifting under generalized closure]\label{thm:lift}
Under Definition~\ref{def:genclosure}, let $h_1$ be a relative certificate on $\Zv$. For
every $\theta$ and every $x$ with $z_\theta(x)\in\Zv$, $h_\theta=h_1\circ z_\theta$ satisfies
(i) $h_\theta(x)<0$ if $x\in\Ob_\theta$, and (ii) \eqref{eq:cbf} at $x$ if $-\delta<h_\theta(x)<\eta$.
\end{theorem}

\begin{proof}
Write $z=z_\theta(x)$. Statement (i) is (V1), since $x\in\Ob_\theta$ gives $s(z)<m_g$. For (ii), the
chain rule and \eqref{eq:genclosure} give $\Lf h_\theta+\Lg h_\theta u=\nabla h_1(z)^\top
F(z,u,\xi^\star)$ for every $u$, with the same $\xi^\star$. Hence
$\max_u[\Lf h_\theta+\Lg h_\theta u]\ge\min_{\xi\in\Xi(z)}\max_u\nabla h_1^\top F(z,u,\xi)\ge-\gamma
h_1(z)$ by (V2).
\end{proof}

The content lies in the hypothesis: $F$ depends on $(z,u,\xi)$ only, so the obstacle, the other
obstacles and the absolute state enter through $z$ alone, and one check over $\Zv$ serves every
obstacle. Condition (V2) is a robust barrier condition with
the residual as the uncertainty. What is specific here is that the range of the residual is computed
from $z$ alone.

\begin{corollary}[Safety until domain exit]\label{cor:safety}
For a map $\{\theta_i\}$ write $h_i=h_{\theta_i}$ and $s_i$ for the signed distances. Call $x$ admissible for obstacle $i$ if $z_{\theta_i}(x)$ lies in the interior of $\Zv$ relative to
$\Zc$ whenever
$-\delta<h_i(x)<\eta$ or $s_i(x)\le m_g$. Let a locally Lipschitz feedback $k$ with values in $U$ satisfy
\begin{equation}\label{eq:joint}
\Lf h_i(x)+\Lg h_i(x)\,k(x)\ \ge\ -\gamma h_i(x)
\end{equation}
for every $i$ and every admissible $x$ in the band of $h_i$, and let $h_i(x(0))\ge0$. If $t_e$ is the first time at which $x(t)$ is not admissible for some obstacle, then
$h_i(x(t))\ge0$ and $x(t)\notin\cup_i\Ob_{\theta_i}$ for all $t\in[0,t_e)$.
\end{corollary}

\begin{proof}
Fix $i$, let $y_i(t)=h_i(x(t))$, and suppose $y_i<0$ somewhere before $t_e$. Let
$t_2=\inf\{t:y_i(t)<0\}$, so $y_i(t_2)=0$, and let $t_1=\sup\{t<t_2:y_i(t)\ge\eta\}$, or $0$ if this set is empty. By continuity $-\delta<y_i<\eta$ on $(t_1,t_2+\epsilon]$ for some
$\epsilon>0$ with $t_2+\epsilon<t_e$, so \eqref{eq:joint} gives $\dot y_i\ge-\gamma y_i$ there,
and the comparison lemma gives
$y_i(t)\ge y_i(t_1)e^{-\gamma(t-t_1)}\ge0$, contradicting the definition of $t_2$. If
$x(t)\in\Ob_{\theta_i}$ for $t<t_e$, admissibility and Theorem~\ref{thm:lift}(i) give
$y_i(t)<0$, a contradiction.
\end{proof}

The guarantee stops at the exit time $t_e$, so keeping the state in the lifted verified domain is part
of the argument. Condition \eqref{eq:joint} asks for one input satisfying all obstacle constraints,
the compatibility question of \cite{breeden2023compositions}, which we assume.

\subsection{When does closure hold?}

\begin{lemma}[Symmetry gives closure]\label{lem:symmetry}
Let a Lie group $G$ act on $\X\times\Theta$ by diffeomorphisms $\Phi_\sigma$ preserving $\D$, and let
$V_u=(f+gu,0)$. If (a) $D\Phi_\sigma V_u=V_u\circ\Phi_\sigma$, (b) $z\circ\Phi_\sigma=z$, (c)
$z(x,\theta)=z(x',\theta')$ implies $(x',\theta')\in G\cdot(x,\theta)$, and (d) $z$ is a submersion
onto a $C^1$ manifold, then Definition~\ref{def:closure} holds.
\end{lemma}

\begin{proof}
Define $F(\zeta,u)=Dz\,V_u$ at any point of $z^{-1}(\zeta)$. By (a) and (b),
$Dz(\Phi_\sigma y)V_u(\Phi_\sigma y)=D(z\circ\Phi_\sigma)(y)V_u(y)=Dz(y)V_u(y)$, and by (c) every point
of the fibre is some $\Phi_\sigma y$, so $F$ is well defined. It is continuous by local sections of
(d), and $Dz\,V_u=D_xz_\theta(f+gu)$ since the $\theta$-component of $V_u$ vanishes.
\end{proof}

This is the classical reduction of a symmetric control system \cite{grizzle1985symmetries}, written
for relative coordinates, and hypothesis (a) is the state-space symmetry of \cite{grizzle1985symmetries} and the
equivariance of \cite{wiltz2025equivariances} with the input map taken as the identity. Hypothesis (a) requires the group to leave the inputs unchanged. When it
moves them and $U$ is not invariant, the orientation of $U$ becomes a residual.

\begin{proposition}[Rank condition]\label{prop:count}
Let $G$ act freely and properly on an open $\D'\subseteq\D$, so that $\D'/G$ has dimension
\begin{equation}\label{eq:count}
N=n+p-\dim G ,
\end{equation}
and let $z$ be $G$-invariant. (i) If $\operatorname{rank}Dz<N$ on an open subset, $z$ does not
separate orbits there. (ii) Where $z$ does not separate orbits, exact closure holds if and only if
$D_xz_\theta(f+gu)$ is constant on each fibre of $z$ and the induced map is continuous. (iii) Rank $N$ does not exclude an ambiguity that leaves a residual.
\end{proposition}

\begin{proof}
For (i), $z=\hat z\circ\pi$ with $\pi$ the quotient map. On the open set where $\hat z$ attains its
largest rank $k<N$ the rank is constant, so $\hat z$ is locally a projection and not injective.
Statement (ii) restates Definition~\ref{def:closure} on fibres, extending to the fibres of $z$ the
requirement, automatic for group orbits \cite{grizzle1985symmetries}, that the vector field project
to a well-defined quotient system, and (iii) is realized in
Section~\ref{sec:pvtol}.
\end{proof}

When the rank falls short of $N$, closure is not implied by symmetry and must be checked. It can be
refuted by two states with equal $z$ and different $\dot z$, found by perturbing the state along the
fibre of $z$.

\begin{lemma}[Pointwise statements need only symmetry]\label{lem:pointwise}
Under (a) and (b) of Lemma~\ref{lem:symmetry}, the margin $\mu(x,\theta)=\max_u[\Lf h_\theta+\Lg
h_\theta u]+\gamma h_\theta$ is $G$-invariant, so the fraction of states satisfying \eqref{eq:cbf}
under any sampling distribution is unchanged by the action of $G$.
\end{lemma}

\begin{proof}
With $H=h_1\circ z$, (b) gives $DH(\Phi_\sigma y)D\Phi_\sigma(y)=DH(y)$, and (a) gives
$DH(\Phi_\sigma y)V_u(\Phi_\sigma y)=DH(y)V_u(y)$ for every $u$.
\end{proof}

A statistical statement at sampled full states therefore transfers across placements without closure,
while a statement over a region of relative coordinates, which Theorem~\ref{thm:lift} consumes, needs
closure or generalized closure.

%%%%%%%%%%%%%%%%%%%%%%%%%%%%%%%%%%%%%%%%%%%%%%%%%%%%%%%%%%%%%%%%%%%%%%%%%%%%%%%%
\section{THREE SYSTEMS}\label{sec:systems}

Write $\rho=s+r$, $\er=(Px-c)/\rho$, $v_r=\vv\cdot\er$, $\ww=\vv-v_r\er$ and $\vperp=\lVert\ww\rVert^2$. All identities below were checked against the true vector fields by finite
differences to within $3\times10^{-5}$. Table~\ref{tab:count} summarizes the counts. The planar residuals are deliberate: the bearing and the signed tangential speed would close them
exactly. We keep them to show that such coordinates remain sound once the residual enters (V2) as a
minimum.

\begin{table}[t]
\caption{Symmetry, count and residual: state dimension $n$, obstacle dimension $p$, symmetry group
$G$, $N=n+p-\dim G$.}
\label{tab:count}
\centering
\setlength{\tabcolsep}{3pt}
\begin{tabular}{@{}lccccl@{}}
\toprule
System & $n$ & $p$ & $\dim G$ & $N$ & rank $z$, residual $\Xi$\\
\midrule
Double integrator & 4 & 3 & 2 & 5 & 4, $S^1\times\{\pm1\}$\\
Planar VTOL, \eqref{eq:zpvtol} & 6 & 3 & 2 & 7 & 7, $\{\pm1\}$\\
Planar VTOL, bearing-blind & 6 & 3 & 2 & 7 & 6, bearing\\
Quadrotor, \eqref{eq:zquad} & 12 & 4 & 4 & 12 & 9, branches, $\hat\omega$\\
Quadrotor, \eqref{eq:zbody} & 12 & 4 & 4 & 12 & 12, singleton\\
\bottomrule
\end{tabular}
\end{table}

\subsection{Double integrator}\label{sec:di}
With $\dot p=\vv$, $\dot\vv=u\in[-a_{\max},a_{\max}]^2$, the box breaks rotations, so $G=\R^2$ and
$N=5$. The coordinates $z=(s,v_r,\vperp,r)$ have rank $4$, with $\dot s=v_r$,
$\dot v_r=u\cdot\er+\vperp/\rho$ and $\tfrac{d}{dt}\vperp=2u\cdot\ww-2v_r\vperp/\rho$. The residual is the
angle $\alpha$ of $\er$ and the sign of $\ww$, which enter only through the input.

\begin{proposition}\label{prop:di}
For the double integrator, with $D(z)=\partial_sh_1v_r+\partial_{v_r}h_1\vperp/\rho-\partial_{\vperp}h_1\,2v_r\vperp/\rho$,
\begin{equation}\label{eq:psidi}
\Psi(z)=D(z)+a_{\max}\sqrt{(\partial_{v_r}h_1)^2+4\vperp(\partial_{\vperp}h_1)^2}+\gamma h_1 .
\end{equation}
\end{proposition}

\begin{proof}
The input term is $\mathbf c^\top R(-\alpha)u$ with
$\mathbf c=(\partial_{v_r}h_1,\pm2\sqrt{\vperp}\partial_{\vperp}h_1)$, whose maximum over the box is
$a_{\max}\lVert R(\alpha)\mathbf c\rVert_1\ge a_{\max}\lVert\mathbf c\rVert_2$, with equality when
$R(\alpha)\mathbf c$ is aligned with an axis.
\end{proof}

\subsection{Planar aircraft}\label{sec:pvtol}
The dynamics are $\dot p=\vv$, $\dot\vv=T\bb-(0,g)$, $\dot\phi=\omega$, $\dot\omega=\tau$ with
$\bb=(-\sin\phi,\cos\phi)$, $T\in[0,T_{\max}]$ and $|\tau|\le\tau_{\max}$. Gravity breaks rotations,
so $G=\R^2$ and $N=7$. With $\psi$ the angle from $\er$ to $\bb$ and $w=\pm\sqrt{\vperp}$ the signed
tangential speed, we use
\begin{equation}\label{eq:zpvtol}
z=(s,\ v_r,\ \vperp,\ r,\ \cos\psi,\ \sin\psi,\ \omega,\ e_{r,1},\ e_{r,2}),
\end{equation}
of rank $7$. Its derivative is $\dot v_r=T\cos\psi-ge_{r,2}+\vperp/\rho$,
$\tfrac{d}{dt}\vperp=2w(T\sin\psi-ge_{r,1})-2v_r\vperp/\rho$, $\dot\psi=\omega-w/\rho$, and
$\dot\er=(w/\rho)\mathbf e_\perp$. It depends on $z$, $u$ and the sign of $w$ only, so $\Xi=\{\pm1\}$,
and with $c_T=\partial_{v_r}h_1\cos\psi+2w\partial_{\vperp}h_1\sin\psi$
\begin{equation}\label{eq:psipvtol}
\begin{split}
\Psi(z)=\min_{w=\pm\sqrt{\vperp}}\big[&D(z,w)+T_{\max}\max(0,c_T)\\
&+\tau_{\max}|\partial_\omega h_1|\big]+\gamma h_1,
\end{split}
\end{equation}
where $D(z,w)$ collects the input-free terms.

\begin{proposition}[Bearing-blind coordinates]\label{prop:bearing}
$z'=(s,v_r,\vperp,r,\cos\psi,\sin\psi,\omega)$ has rank $6<N$ and does not close.
\end{proposition}

\begin{proof}
A rigid rotation of the configuration about the obstacle centre preserves $z'$ but changes the
bearing, and by the equations above changes $\dot v_r$ by $-g\Delta e_{r,2}\neq0$ for the same input.
\end{proof}

A vehicle approaching an obstacle from below can brake only by cutting thrust, at most at $g$, while one
approaching from above brakes at up to $T_{\max}-g$, and a barrier that cannot see the bearing cannot
tell the two apart (Section~\ref{sec:negative}).

\subsection{Quadrotor}\label{sec:quad}
The state is $(p,\vv,\Theta,\omega)$ with $\dot\vv=T\bb-g\evert$, $\bb=R(\Theta)\evert$, body rates $\omega$ with $\dot\omega=\tau$, a unit-inertia model (a full $J\dot\omega=\tau-\omega\times J\omega$ would not change the closure below), $T\in[0,T_{\max}]$ and $\lVert\tau\rVert_\infty\le\tau_{\max}$.
Translations and rotations about the vertical are symmetries, so $N=12+4-4=12$.

\paragraph*{Nine yaw-invariant scalars}
The coordinates
\begin{equation}\label{eq:zquad}
z=(s,\ v_r,\ \vperp,\ r,\ b_r,\ b_3,\ e_{r,3},\ \beta,\ \lVert\omega\rVert),
\end{equation}
with $b_r=\bb\cdot\er$, $b_3=\bb\cdot\evert$, $e_{r,3}=\er\cdot\evert$ and $\beta=\bb\cdot\ww$, have
rank $9$. They miss the rotation of the body about $\bb$ and the direction of $\omega$.

\begin{proposition}\label{prop:quadfail}
Exact closure holds for $s$, $v_r$ and $r$ in \eqref{eq:zquad} and fails for the other six components.
\end{proposition}

\begin{proof}
Reflecting $\ww$ about $\bb$ in the tangential plane preserves $z$ but changes $\evert\cdot\ww$, which
enters $\tfrac{d}{dt}\vperp$, $\dot e_{r,3}$ and $\dot\beta$. Rotating $\omega$ about $\bb$ preserves $z$
but changes $(R\omega)\times\bb$ and $\hat\omega$, which enter $\dot b_r$, $\dot b_3$ and
$\tfrac{d}{dt}\lVert\omega\rVert=\hat\omega\cdot\tau$.
\end{proof}

Generalized closure holds with at most four discrete branches for $(\bb,\ww)$, fixed by $z$ up to
reflections, and the direction of $\omega$ as a continuous residual, which gives a sound bound on $\Psi$ computable from $z$. We use \eqref{eq:zquad} as a diagnostic,
and the deployed results use \eqref{eq:zbody}. The component $\lVert\omega\rVert$ is, however, not differentiable at $\omega=0$, so
\eqref{eq:zquad} violates the standing smoothness assumption at hover.\label{rem:kink} Since the torque
enters $\dot z$ only through $\hat\omega\cdot\tau$, at $\omega\neq0$ it adds
$\tau_{\max}\lVert\hat\omega\rVert_1|\partial h_1/\partial\lVert\omega\rVert|$ to the margin, the
authority to reduce the rate, which vanishes at $\omega=0$, so a condition checked at $\omega\neq0$
overstates the margin near hover.

\paragraph*{Body-frame coordinates}
With $R$ the body-to-world rotation, $\mathbf d=p-c$, $\mathbf n=R^\top\mathbf d/\rho$,
$\vv_B=R^\top\vv$ and $\mathbf g_B=R^\top\evert$, let
\begin{equation}\label{eq:zbody}
z_B=(s,\ v_r,\ g_r,\ r,\ \mathbf n,\ \vv_B,\ \mathbf g_B,\ \omega)\in\R^{16},
\end{equation}
where $v_r=\mathbf n\cdot\vv_B$ and $g_r=\mathbf n\cdot\mathbf g_B$. Geometric quadrotor control writes $\omega$ in the body frame \cite{lee2010geometric}, and $z_B$ writes
the relative position, the velocity and gravity there too.

\begin{lemma}\label{lem:body}
Along the quadrotor dynamics, $\dot s=v_r$, $\dot r=0$, $\dot\omega=\tau$,
$\dot v_r=(\lVert\vv_B\rVert^2-v_r^2)/\rho+Tn_3-gg_r$, $\dot g_r=(\vv_B\cdot\mathbf g_B-v_rg_r)/\rho$,
$\dot{\mathbf n}=-\omega\times\mathbf n+(\vv_B-v_r\mathbf n)/\rho$,
$\dot\vv_B=-\omega\times\vv_B+T\evert-g\mathbf g_B$ and $\dot{\mathbf g}_B=-\omega\times\mathbf g_B$.
\end{lemma}

\begin{proof}
For a world vector $\mathbf y$, $\tfrac{d}{dt}R^\top\mathbf y=-\omega\times R^\top\mathbf y+R^\top\dot{\mathbf y}$.
Apply this to $\mathbf d$, $\vv$ and $\evert$, use $\dot\rho=v_r$, and note that the $\omega$ terms
cancel in $v_r$ and $g_r$.
\end{proof}

\begin{proposition}[Body-frame coordinates close]\label{prop:body}
The map \eqref{eq:zbody} is $G$-invariant, smooth on $\D$, has rank $12=N$, and closes exactly with $F$
given by Lemma~\ref{lem:body}.
\end{proposition}

\begin{proof}
A group element maps $(\mathbf d,\vv,R)$ to $(R_\alpha\mathbf d,R_\alpha\vv,R_\alpha R)$ with $R_\alpha$ a
rotation about $\evert$ and leaves $\omega$ unchanged, so $z_B$ is invariant. Conversely $\mathbf g_B$
fixes $R$ up to such a rotation, and then $\mathbf d$ and $\vv$ up to the same rotation, so the fibres
are the four-dimensional orbits and the rank is $n+p-4=12$. Closure is Lemma~\ref{lem:body}.
\end{proof}

With exact closure, (V2) becomes $\Psi(z_B)=\nabla h_1^\top F(z_B,0)+T_{\max}\max(0,c_T)+\tau_{\max}\lVert\nabla_\omega h_1\rVert_1+\gamma h_1$ with $c_T=n_3\partial_{v_r}h_1+\partial_{v_{B,3}}h_1$, and the torque now enters
through $\dot\omega=\tau$ directly.

%%%%%%%%%%%%%%%%%%%%%%%%%%%%%%%%%%%%%%%%%%%%%%%%%%%%%%%%%%%%%%%%%%%%%%%%%%%%%%%%
\section{LEARNING, VERIFICATION AND DEPLOYMENT}\label{sec:learning}

\paragraph*{Training}
Following the practice of learned certificates \cite{dawson2023survey}, $h_1$ is a $\tanh$ multilayer perceptron with three hidden layers of width $96$ (double integrator) or $128$,
so it is smooth. At each epoch a map with one to four spheres is sampled, and three one-sided penalties are applied to the composite $h=-\kappa^{-1}\log\sum_ie^{-\kappa h_1(z_i)}$
with $\kappa=8$: classification of labelled states, the descent condition \eqref{eq:cbf} with the
exact box optimum, and the discrete decrease $h(x^+)-h(x)\ge-\gamma h\,dt$ after one step from a
filtered nominal input. The composite is a smooth approximation of $\min_ih_1(z_i)$, which is a nonsmooth barrier whenever one input meets every active component condition
\cite{glotfelter2017nonsmooth}. (V2) and deployment use the per-obstacle barriers directly.

\paragraph*{Labels}
A state is labelled unsafe if the braking policy, simulated from it with the known dynamics over a
horizon long enough to stop, enters the inflated obstacle, and safe otherwise.
The labels are conservative, since a state from which an admissible feedback avoids $\Ob_\theta$ for
all time lies in the viability kernel of its complement \cite{aubin1991viability,gurriet2018backup}.

For the double integrator, where the discrete kernel \cite{aubin1991viability} can be computed on
an $81^3$ grid at fixed radius, the labels recover $99.28\%$ of it. For the quadrotor the labelling policy is the deployed brake together with the
confinement filter that deployment applies to it, which keeps roll and pitch within $0.475$\,rad,
simulated for $10.2$\,s. The quadrotor barrier is trained in two stages: $5000$ epochs with labels
from the unfiltered brake, then $2000$ epochs at half the learning rate with the filtered brake, so
that the labels describe the brake as deployed, and with a quarter of the states drawn from fast
approaches tilted toward an obstacle, which are rare under uniform sampling.

\paragraph*{Verification}
For a box $B$ of relative coordinates, certified CROWN bound propagation
\cite{zhang2018crown},
as in \cite{vertovec2025scalable}, encloses $h_1$ and $\nabla h_1$, and with \eqref{eq:psidi} or \eqref{eq:psipvtol} gives a lower bound on
$\Psi$ over $B$. A nonnegative bound certifies (V2) on all of $B$, and a negative upper bound on $h_1$
certifies (V1). We draw the positions of boxes of side $\lambda$ independently and uniformly over $\Zv$ and report the
certified fraction with a
one-sided Clopper--Pearson bound \cite{clopper1934}.
Every bound behind Table~\ref{tab:verif}, computed in double precision, clears its threshold by at
least $1.3\times10^{-5}$. For the quadrotor we check \eqref{eq:cbf} at sampled
full states with exact Lie derivatives, which transfers by Lemma~\ref{lem:pointwise}, and for
\eqref{eq:zbody} also on the relative coordinates, where exact closure makes the two coincide.
These checks are pointwise, so for the quadrotor Theorem~\ref{thm:lift} is not exercised over
regions of $\Zv$.
\vspace{-1pt}
% \subsection{Comparison with a hand-designed barrier}\label{sec:compare}
\paragraph*{Comparison with a hand-designed barrier}
As baseline we use the second-order high-order control barrier function (HOCBF) on the distance
\cite{xiao2019hocbf}, $h_0=\lVert
Px-c\rVert^2-(r+m_g)^2$ and $\psi_1=\dot h_0+\alpha_1h_0$, with the constraint $\dot\psi_1+\gamma\psi_1\ge0$
in place of the learned one and the controller otherwise unchanged. We report $\alpha_1\in\{0.5,1,2,4\}$, with $\gamma=0.6$ as for the learned barrier, on the held-out
maps and on fast starts, which begin $0.6$ to $2.5$\,m from an
obstacle and move toward it at $1.5$ to $3.5$\,m/s (double integrator) or $1$ to $2.8$\,m/s, within the
velocity ranges of $\Zv$. For each fast start we record whether it lies in the
method's own safe set, $\{h_1\ge0\}$ for the learned barrier and $\{h_0\ge0\}\cap\{\psi_1\ge0\}$ for the
HOCBF, in which case we say the method admits it, and whether braking alone avoids contact from it
(Table~\ref{tab:compare}).

\hspace*{\parindent}On the two underactuated vehicles none of the $118$ fast starts admitted by the learned barrier
collided ($70$ on the planar aircraft, $48$ on the quadrotor with each check). Every start it admits is
recoverable by braking, on all three systems, and it admits $71\%$, $47\%$ and $47\%$ of the
recoverable starts of the double integrator, the planar aircraft and the quadrotor. On the two
vehicles the HOCBF admits few fast starts at low gains, and at the gains where it admits more, $19$ to
$36\%$ of its admitted starts collided and $28$ to $134$ could not be saved by braking ($28$ and $90$
on the planar aircraft, $65$ and $134$ on the quadrotor). In closed loop, almost all learned
fast-start collisions come from starts that braking cannot save ($115$ of $126$ on the planar
aircraft, all on the quadrotor). The high-gain HOCBF collides less from recoverable planar-aircraft
starts ($2$ against $11$ of $148$) and less overall on the quadrotor ($133$ to $164$ against $182$ and
$183$), but penetrates obstacles from rest on the held-out maps, where the learned planar-aircraft
barrier entered the inflation in $4$ runs, by up to $31.5$\,cm, without collision. On the double
integrator the HOCBF at $\alpha_1=4$ is correct and less conservative, and is the better choice. On
the quadrotor the HOCBF at $\alpha_1=1$ with the backup check also had no collision from rest or from
its $27$ admitted fast starts, with higher reach. The check alone does not suffice: with no barrier
constraint in the QP and the check applied to the inflated distance, $10$ and $23$ of $288$ held-out
runs from rest collided with the backup and one-step checks (one-sided Fisher $p<0.001$ against $0$).
A barrier constraint is thus needed, the learned one and the HOCBF at $\alpha_1=1$ with the backup
check were equally safe in these tests, and the learned barrier's advantage lies in which states it
admits.

\paragraph*{Deployment}
At each step the controller solves the CBF quadratic program \cite{ames2017cbfqp} in its
safety-filter form \cite{gurriet2018backup}
\begin{equation}\label{eq:qp}
\begin{aligned}
\min_{u\in U,\nu\ge0}\ &\lVert u-u_{\mathrm{nom}}\rVert^2+\varpi\,\mathbf 1^\top\nu\\
\text{s.t. }&\Lf h_i+\Lg h_iu\ge-\gamma h_i,\quad i=1,\dots,K,\\
&c_j+a_ju\ge-\nu_j,\quad j=1,\dots,M,
\end{aligned}
\end{equation}
with one hard constraint per obstacle, which enforces the intersection of the per-obstacle safe sets
\cite{glotfelter2017nonsmooth}, and soft confinement constraints with a linear penalty $\varpi=10^4$, which enforces them exactly
when they can be met and $\varpi$ exceeds their multipliers. The nominal input is a cascade in the style of \cite{lee2010geometric}.
For the quadrotor the confinement constrains roll and pitch to $0.5$\,rad through exact second-order
constraints in the manner of \cite{xiao2019hocbf}, with $[\ddot\phi,\ddot\vartheta]$ affine in
$\tau$, and velocity and rates within their training ranges. The verification samples roll, pitch and yaw
in $[-0.5,0.5]$, $|v_i|\le2$, $|\omega_i|\le1.5$ and obstacle offsets in $[-5,5]^2\times[-3,3]$, and a rotation about the vertical maps every state with $|\phi|,|\vartheta|\le0.5$,
$|\omega_i|\le1.5$, $|v_3|\le2$, horizontal speed at most $2$, vertical offset within $3$\,m and
horizontal offset at most $5$\,m, at any yaw, onto a sampled one.

The input is held over steps of $0.05$\,s. The candidate is applied to the known dynamics for one step,
followed by $K_b$ braking steps, and accepted if $\min_ih_i\ge0$ along the horizon. Otherwise the QP is re-solved with each obstacle constraint required to hold with a margin of
$0.02$, $0.05$, $0.1$ and $0.2$ in turn, then the
max-margin input, which maximizes $\min_i(\Lf h_i+\Lg h_iu+\gamma h_i)$ over $U$, is used
if it passes the same check, and otherwise the brake, which plays the role of the backup policy of backup control barrier
functions \cite{chen2021backup}. We report $K_b=0$ and
$K_b=26$. None of these discrete-time devices is certified. The planar systems use hard confinement
and a one-step check.

%%%%%%%%%%%%%%%%%%%%%%%%%%%%%%%%%%%%%%%%%%%%%%%%%%%%%%%%%%%%%%%%%%%%%%%%%%%%%%%%
\section{RESULTS}\label{sec:results}

All systems (DI: double integrator, VTOL: planar aircraft, Quad: quadrotor) use $dt=0.05$\,s, a $30$\,s horizon, $m_g=0.4$, $\gamma=0.6$ and $\kappa=8$.  Each evaluation
uses $24$ maps with one to four spheres of radius in $[0.6,1.5]$ (quadrotor $[0.6,1.4]$) and $12$
starts at rest per map. These maps were used while the controller was developed, and $24$ further
held-out maps were not. The trained network is
re-pointed to the obstacles of each map without retraining. Reach is the fraction of runs that come
within $0.4$\,m of the goal ($0.5$\,m for the quadrotor) inside the horizon. A collision is contact
with a true, uninflated obstacle. The minimum distance is the smallest signed distance to a true
surface, so a value below $m_g$ means that the inflated obstacle, which the barrier is meant to
exclude, was entered.

\subsection{Verification}
Table~\ref{tab:verif} reports the planar results. The relative domain for (V2) is $s\in[0.4,4]$,
$v_r\in[-3.54,3.54]$, $\vperp\in[0,12.5]$, $r\in[0.6,1.5]$ for the double integrator, and the same
ranges of $s$ and $r$ with $v_r\in[-2.83,2.83]$, $\vperp\in[0,8]$, $\omega\in[-1.5,1.5]$ and unrestricted angles for the planar aircraft. (V1) is checked on $s\in[-0.5,0.35]$ over the same ranges and the band form of (V2) on
$s\in[0.35,4]$, with $\delta=0.02$ and $\eta=0.05$. The two checks meet, so the analysis uses the
inflation $0.35$\,m in (V1) and Corollary~\ref{cor:safety}, slightly inside the $0.4$\,m used for
training and control, and the checked domain has no gap in $s$. Where some $h_i$ first reaches zero, (V1) places the state at
$s>0.35$, so the states that the proof of Corollary~\ref{cor:safety} uses lie in the range of the band
check. The certified (V2) fraction of the double integrator
rises from $35.1\%$ to $99.6\%$ as the box side shrinks from $0.1$ to $0.00125$, against $99.956\%$ of
sampled points at which (V2) holds, which bounds what any box method can certify.  Beyond $s=4$ every double-integrator box has
$h_1\ge0.05$, so the band cannot occur there, while for the planar aircraft $0.59\%$ of far-field
boxes are not certified above the band, so there the band hypothesis holds only for $s\le4$.

\begin{table}[t]
\caption{Box verification: certified fraction of sampled boxes [one-sided $95\%$ Clopper--Pearson
lower bound]. $n$: boxes, of $2{,}500$ drawn, that meet the set checked. On band: (V2) on
$-0.02<h_1<0.05$ over $s\in[0.35,4]$, same box side.}
\label{tab:verif}
\centering
\setlength{\tabcolsep}{3pt}
\begin{tabular}{@{}llcc@{}}
\toprule
System & Statement & $n$ & Fraction [CP]\\
\midrule
DI & (V2), side $1.25\times10^{-3}$ & $1583$ & $99.62\%$ [$99.25\%$]\\
DI & (V1), side $1.25\times10^{-3}$ & $1664$ & $100\%$ [$99.82\%$]\\
DI & (V2) on band & $1564$ & $99.36\%$ [$98.91\%$]\\
VTOL & (V2), side $3.1\times10^{-4}$ & $1431$ & $97.20\%$ [$96.37\%$]\\
VTOL & (V1), side $0.01$ & $1692$ & $99.65\%$ [$99.30\%$]\\
VTOL & (V2) on band & $1429$ & $96.85\%$ [$95.98\%$]\\
\bottomrule
\end{tabular}
\end{table}

\subsection{The bearing-blind coordinates}\label{sec:negative}
Planar-aircraft barriers trained identically, but on the bearing-blind coordinates of
Proposition~\ref{prop:bearing} or on \eqref{eq:zpvtol}, were compared twice. From rest, without the workspace constraints that keep the position inside the map, the bearing-blind barrier penetrated an obstacle in $3$ of $288$ deployments,
to $-0.023$\,m, and \eqref{eq:zpvtol} in none, at unchanged reach ($77.8$ and $78.8\%$, against $97.2\%$ with the workspace constraints of
Table~\ref{tab:deploy}), a difference
too small to be significant on its own. The proposition says where to look, so the second test, with
the deployed controller and a bearing-blind barrier retrained by the same recipe, started both from
$288$ fast starts within $45^\circ$ of directly below an obstacle and $288$ from directly above. From
below, the bearing-blind barrier collided from $16$ of the $65$ starts that braking alone can save,
against $5$ for \eqref{eq:zpvtol} (one-sided Fisher exact $p=0.008$), and reached the goal less often
($71$ against $93$ runs). From above, where braking authority is larger, the two are
indistinguishable ($4$ and $5$ of $203$ recoverable starts). The bearing-blind coordinates still close up to a residual, the bearing, so
Theorem~\ref{thm:lift} would apply if (V2) took the minimum over every bearing, which forces the
barrier to assume the weaker braking from below in every direction. The trained barrier ignores the
residual and fails from below, where it matters.

\subsection{The quadrotor}
Table~\ref{tab:quadverif} reports the pointwise checks on the band. Away from hover both barriers meet the condition on at least $99.9\%$ of it. At hover the barrier in \eqref{eq:zquad} meets it on only
$69.6\%$, as Section~\ref{sec:quad} predicts, while the body-frame barrier meets it on $99.1\%$. For
\eqref{eq:zbody} the relative-coordinate formula agreed with the full-state margin to
$8.1\times10^{-7}$, as exact closure requires, and classification (V1) failed at $55$ of $4.4\times10^5$ sampled states, all with $s\ge0.349$, only $1$ of them inside the inflation $0.35$\,m of the analysis, and none inside
the true obstacle.

\begin{table*}[t]
\caption{Quadrotor, fraction of sampled band states meeting \eqref{eq:cbf} [CP $95\%$ lower bound],
from $2{,}018$ to $6{,}200$ band states per cell. At $\omega=0$, \eqref{eq:zquad} uses its torque-free margin.}
\label{tab:quadverif}
\centering
\setlength{\tabcolsep}{5pt}
\begin{tabular}{@{}llcccc@{}}
\toprule
Coordinates & Statement & $r=0.6$ & $r=1.0$ & $r=1.4$ & $r\sim U[0.6,1.4]$\\
\midrule
\eqref{eq:zquad}, rank $9$ & true state & $100\%$ [$99.88\%$] & $99.94\%$ [$99.79\%$] & $99.95\%$ [$99.83\%$] & $99.97\%$ [$99.89\%$]\\
& hover $\omega=0$ & & & & $69.61\%$ [$68.39\%$]\\
\eqref{eq:zbody}, rank $12$ & true state & $99.90\%$ [$99.68\%$] & $100\%$ [$99.88\%$] & $99.90\%$ [$99.74\%$] & $99.98\%$ [$99.90\%$]\\
& hover $\omega=0$ & & & & $99.06\%$ [$98.80\%$]\\
\bottomrule
\end{tabular}
\end{table*}

\begin{table*}[t]
\caption{Learned barrier against the distance HOCBF in the same controller, $288$ deployments per
cell. Admitted: fast starts in the method's own safe set. Coll.\ from adm., from recov.: collisions from
admitted starts, and from starts that braking alone can save, whose number differs by system.}
\label{tab:compare}
\centering
\setlength{\tabcolsep}{4pt}
\begin{tabular}{@{}llccccccc@{}}
\toprule
& & \multicolumn{3}{c}{Held-out, at rest} & \multicolumn{4}{c}{Fast starts}\\
\cmidrule(lr){3-5}\cmidrule(l){6-9}
System & Barrier & Coll. & Reach & Min.\ dist. & Coll. & Admitted & Coll.\ from adm. & Coll.\ from recov.\\
\midrule
DI & learned & $0/288$ & $97.9\%$ & $0.386$ & $47/288$ & $187$ & $0$ & $22/262$\\
DI & HOCBF, $\alpha_1=0.5$ & $0/288$ & $97.6\%$ & $0.400$ & $101/288$ & $0$ & $0$ & $76/262$\\
DI & HOCBF, $\alpha_1=1$ & $0/288$ & $97.6\%$ & $0.400$ & $86/288$ & $1$ & $0$ & $61/262$\\
DI & HOCBF, $\alpha_1=2$ & $0/288$ & $97.6\%$ & $0.400$ & $52/288$ & $80$ & $0$ & $27/262$\\
DI & HOCBF, $\alpha_1=4$ & $0/288$ & $97.6\%$ & $0.400$ & $24/288$ & $205$ & $0$ & $0/262$\\
\midrule
VTOL & learned & $0/288$ & $97.6\%$ & $0.085$ & $126/288$ & $70$ & $0$ & $11/148$\\
VTOL & HOCBF, $\alpha_1=0.5$ & $0/288$ & $88.5\%$ & $0.104$ & $153/288$ & $0$ & $0$ & $38/148$\\
VTOL & HOCBF, $\alpha_1=1$ & $0/288$ & $92.4\%$ & $0.065$ & $128/288$ & $24$ & $0$ & $14/148$\\
VTOL & HOCBF, $\alpha_1=2$ & $1/288$ & $91.7\%$ & $-0.006$ & $131/288$ & $138$ & $26$ & $2/148$\\
VTOL & HOCBF, $\alpha_1=4$ & $3/288$ & $90.6\%$ & $-0.009$ & $133/288$ & $235$ & $84$ & $2/148$\\
\midrule
Quad & learned, one-step & $0/288$ & $99.0\%$ & $0.150$ & $182/288$ & $48$ & $0$ & $0/103$\\
Quad & learned, backup & $0/288$ & $97.2\%$ & $0.253$ & $183/288$ & $48$ & $0$ & $0/103$\\
Quad & HOCBF, $\alpha_1=0.5$, one-step & $0/288$ & $99.7\%$ & $0.192$ & $183/288$ & $0$ & $0$ & $0/103$\\
Quad & HOCBF, $\alpha_1=1$, one-step & $1/288$ & $98.6\%$ & $>-0.001$ & $184/288$ & $27$ & $1$ & $1/103$\\
Quad & HOCBF, $\alpha_1=2$, one-step & $2/288$ & $97.9\%$ & $-0.017$ & $164/288$ & $151$ & $46$ & $1/103$\\
Quad & HOCBF, $\alpha_1=4$, one-step & $2/288$ & $98.3\%$ & $-0.005$ & $133/288$ & $234$ & $83$ & $1/103$\\
Quad & check only, one-step & $23/288$ & $91.7\%$ & $-0.054$ & $199/288$ & -- & -- & $26/103$\\
Quad & check only, backup & $10/288$ & $88.9\%$ & $-0.041$ & $190/288$ & -- & -- & $7/103$\\
Quad & HOCBF, $\alpha_1=1$, backup & $0/288$ & $99.3\%$ & $0.414$ & $183/288$ & $27$ & $0$ & $0/103$\\
\bottomrule
\end{tabular}
\end{table*}

\begin{figure}[t]
\centering
\includegraphics[width=0.65\columnwidth,trim=372 190 0 14,clip]{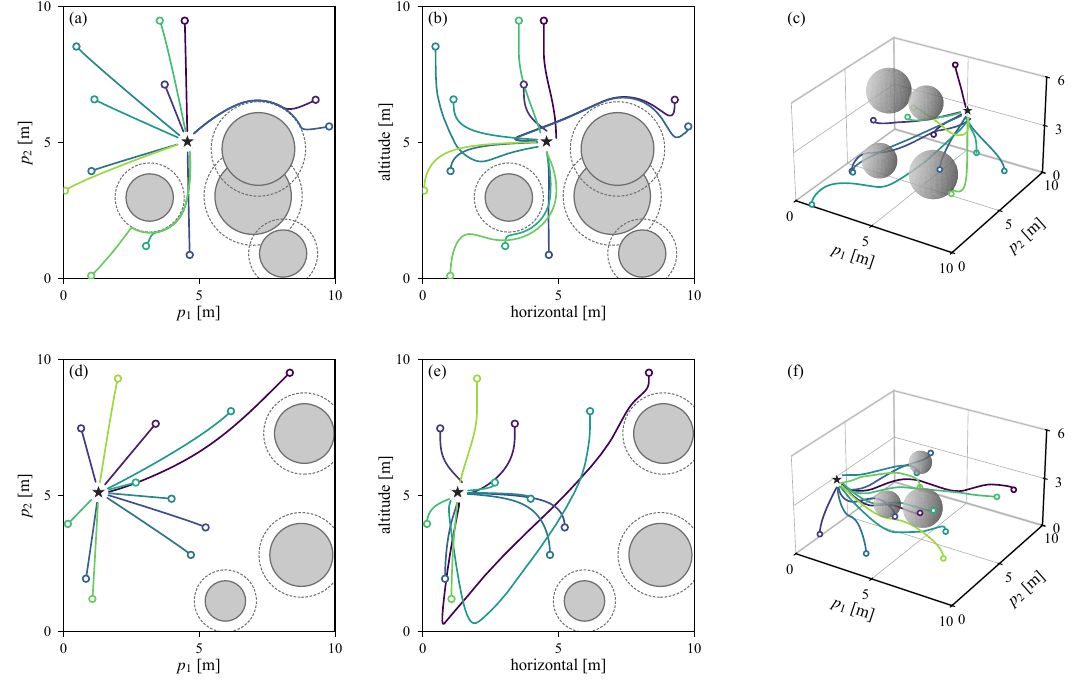}
\caption{Quadrotor with the body-frame barrier and the backup check on the first evaluation map,
not selected. All $12$ starts reach the goal, at least $0.453$\,m from every
obstacle surface}
\label{fig:traj}
\end{figure}

\paragraph*{Deployment}
Table~\ref{tab:deploy} reports the deployments, and Fig.~\ref{fig:traj} shows the quadrotor on one map. From the $576$ quadrotor starts at rest, $288$ on the evaluation maps and $288$ on the held-out
maps, the body-frame barrier had no collision with either check. Along every quadrotor trajectory
(V2) held at every visited band state, and the state left the lifted verified domain once, for $7$
steps, with roll or pitch at $29.6^\circ$, where the obstacle constraints could be met only by
relaxing the soft attitude constraints. The barrier's own specification was violated more often. In
every row some $h_i$ became negative in discrete time, because over one held step $h_1$ can fall
faster than the continuous-time condition allows, and $1$ to $7$ runs per row entered the inflated
obstacle, by up to $25$\,cm. The discrete-time check and the brake, acting at $2$ to $13\%$ of
quadrotor steps, kept every run off the true obstacles. The obstacle constraints of the QP were jointly infeasible at a single time step across all
$1{,}152$ quadrotor deployments, so the fallbacks acted almost always because the QP solution failed
the check. These results therefore rest on uncertified devices as well as on the barrier. With the barrier and the QP alone, without the check, the double
integrator and the planar aircraft still had no collision, and on the quadrotor $5$ of $576$ runs
collided, by at most $6.6$\,cm.

\begin{table}[t]
\caption{Deployments from rest, $288$ per row ($24$ maps, $12$ starts each), quadrotor with
\eqref{eq:zbody}. CP: $95\%$ upper bound on the collision probability of a start, treating the runs as
independent although starts on a map share it. Min.\ dist.: to a true surface, in m. Margin: runs that entered the
inflated obstacle. $h<0$: runs in which some $h_i$ became negative.}
\label{tab:deploy}
\centering
\setlength{\tabcolsep}{2.2pt}
\begin{tabular}{@{}lcccccc@{}}
\toprule
System & Coll. & CP & Min.\ dist. & Margin & $h<0$ & Reach\\
\midrule
DI & $0$ & $1.04\%$ & $0.388$ & $1$ & $16$ & $96.9\%$\\
VTOL, \eqref{eq:zpvtol} & $0$ & $1.04\%$ & $0.236$ & $1$ & $10$ & $97.2\%$\\
Quad, one-step & $0$ & $1.04\%$ & $0.240$ & $4$ & $4$ & $99.3\%$\\
Quad, backup & $0$ & $1.04\%$ & $0.378$ & $1$ & $1$ & $99.7\%$\\
Quad, held-out, one-step & $0$ & $1.04\%$ & $0.150$ & $7$ & $11$ & $99.0\%$\\
Quad, held-out, backup & $0$ & $1.04\%$ & $0.253$ & $6$ & $11$ & $97.2\%$\\
\bottomrule
\end{tabular}
\end{table}

%%%%%%%%%%%%%%%%%%%%%%%%%%%%%%%%%%%%%%%%%%%%%%%%%%%%%%%%%%%%%%%%%%%%%%%%%%%%%%%%
\section{CONCLUSION AND LIMITATIONS}\label{sec:limits}

This paper developed a way to transfer a neural CBF across obstacle
configurations. A generalized closure condition captures the information that relative coordinates
omit, and the lifting theorem shows how a certificate in those coordinates applies to every obstacle
and map. The rank test identifies when symmetry alone is insufficient. Across a double integrator,
a planar aircraft, and a 12-state quadrotor, the experiments show that coordinates failing the
condition can cause penetration or poor behavior near hover, while the body-frame coordinates give
a transferable quadrotor barrier. In the reported deployments, it transferred to held-out maps
without retraining and avoided collision from all tested fast starts that it admitted.

The coverage of the checked set is statistical, and the quadrotor check is pointwise. The theory is
in continuous time, and the discrete-time safeguards are not certified. A sampled-data form of (V2)
as in \cite{breeden2022sampled} would close that gap and will be addressed in future works. On $288$ runs on maps with $6$ to $10$
obstacles and narrow gaps, the quadrotor barrier collided once with the one-step check and never
with the backup check, which reaches the goal less often than the hand-designed barrier. Each
barrier was trained with one seed, joint feasibility is assumed, and obstacles are static spheres.

Future work should develop sampled-data certificates for the deployed safeguards, quantify coverage
over larger and dynamic obstacle families, and study training procedures that provide repeatable
certificates.

%%%%%%%%%%%%%%%%%%%%%%%%%%%%%%%%%%%%%%%%%%%%%%%%%%%%%%%%%%%%%%%%%%%%%%%%%%%%%%%%
% \section*{ACKNOWLEDGMENT}
% Large language model tools assisted with code, analysis and editing, and the author checked all results.

% Reference list inlined from the BibTeX output (IEEEtran.bst) for arXiv.


\begin{thebibliography}{10}
\providecommand{\url}[1]{#1}
\csname url@samestyle\endcsname
\providecommand{\newblock}{\relax}
\providecommand{\bibinfo}[2]{#2}
\providecommand{\BIBentrySTDinterwordspacing}{\spaceskip=0pt\relax}
\providecommand{\BIBentryALTinterwordstretchfactor}{4}
\providecommand{\BIBentryALTinterwordspacing}{\spaceskip=\fontdimen2\font plus
\BIBentryALTinterwordstretchfactor\fontdimen3\font minus
  \fontdimen4\font\relax}
\providecommand{\BIBforeignlanguage}[2]{{%
\expandafter\ifx\csname l@#1\endcsname\relax
\typeout{** WARNING: IEEEtran.bst: No hyphenation pattern has been}%
\typeout{** loaded for the language `#1'. Using the pattern for}%
\typeout{** the default language instead.}%
\else
\language=\csname l@#1\endcsname
\fi
#2}}
\providecommand{\BIBdecl}{\relax}
\BIBdecl

\bibitem{ames2017cbfqp}
A.~D. Ames, X.~Xu, J.~W. Grizzle, and P.~Tabuada, ``Control barrier function
  based quadratic programs for safety critical systems,'' \emph{IEEE
  Transactions on Automatic Control}, vol.~62, no.~8, pp. 3861--3876, 2017.

\bibitem{gurriet2018backup}
T.~Gurriet, A.~Singletary, J.~Reher, L.~Ciarletta, E.~Feron, and A.~Ames,
  ``Towards a framework for realizable safety critical control through active
  set invariance,'' in \emph{2018 ACM/IEEE 9th International Conference on
  Cyber-Physical Systems (ICCPS)}, 2018, pp. 98--106.

\bibitem{xiao2019hocbf}
W.~Xiao and C.~Belta, ``Control barrier functions for systems with high
  relative degree,'' in \emph{2019 IEEE 58th Conference on Decision and Control
  (CDC)}, 2019, pp. 474--479.

\bibitem{aubin1991viability}
J.-P. Aubin, A.~M. Bayen, and P.~Saint-Pierre, \emph{Viability Theory: New
  Directions}, 2nd~ed.\hskip 1em plus 0.5em minus 0.4em\relax Springer Berlin
  Heidelberg, 2011.

\bibitem{bansal2017hj}
S.~Bansal, M.~Chen, S.~Herbert, and C.~J. Tomlin, ``Hamilton-{J}acobi
  reachability: A brief overview and recent advances,'' in \emph{2017 IEEE 56th
  Annual Conference on Decision and Control (CDC)}, 2017, pp. 2242--2253.

\bibitem{bansal2021deepreach}
S.~Bansal and C.~J. Tomlin, ``{DeepReach}: A deep learning approach to
  high-dimensional reachability,'' in \emph{2021 IEEE International Conference
  on Robotics and Automation (ICRA)}, 2021, pp. 1817--1824.

\bibitem{dawson2023survey}
C.~Dawson, S.~Gao, and C.~Fan, ``Safe control with learned certificates: A
  survey of neural {L}yapunov, barrier, and contraction methods for robotics
  and control,'' \emph{IEEE Transactions on Robotics}, vol.~39, no.~3, pp.
  1749--1767, 2023.

\bibitem{zhang2023exact}
H.~Zhang, J.~Wu, Y.~Vorobeychik, and A.~Clark, ``Exact verification of {ReLU}
  neural control barrier functions,'' in \emph{Advances in Neural Information
  Processing Systems (NeurIPS)}, vol.~36, 2023, arXiv:2310.09360.

\bibitem{hu2024symbolic}
H.~Hu, Y.~Yang, T.~Wei, and C.~Liu, ``Verification of neural control barrier
  functions with symbolic derivative bounds propagation,'' in \emph{Proceedings
  of The 8th Conference on Robot Learning}, vol. 270, 2025, pp. 1797--1814.

\bibitem{vertovec2025scalable}
N.~Vertovec, F.~B. Mathiesen, T.~Badings, L.~Laurenti, and A.~Abate, ``Scalable
  verification of neural control barrier functions using linear bound
  propagation,'' in \emph{Learning for Dynamics and Control Conference (L4DC)},
  2026, arXiv:2511.06341.

\bibitem{tayal2024verified}
M.~Tayal, H.~Zhang, P.~Jagtap, A.~Clark, and S.~Kolathaya, ``Learning a
  formally verified control barrier function in stochastic environment,'' in
  \emph{2024 IEEE 63rd Conference on Decision and Control (CDC)}, 2024, pp.
  4098--4104.

\bibitem{qin2021decentralized}
Z.~Qin, K.~Zhang, Y.~Chen, J.~Chen, and C.~Fan, ``Learning safe multi-agent
  control with decentralized neural barrier certificates,'' in
  \emph{International Conference on Learning Representations (ICLR)}, 2021,
  arXiv:2101.05436.

\bibitem{zhang2025gcbfplus}
S.~Zhang, O.~So, K.~Garg, and C.~Fan, ``{GCBF+}: A neural graph control barrier
  function framework for distributed safe multiagent control,'' \emph{IEEE
  Transactions on Robotics}, vol.~41, pp. 1533--1552, 2025.

\bibitem{derajic2026cncbf}
B.~Derajić, S.~Bernhard, and W.~Hönig, ``{CN-CBF}: Composite neural control
  barrier function for robot navigation in dynamic environments,'' 2026,
  arXiv:2603.06921.

\bibitem{wiltz2025equivariances}
A.~Wiltz and D.~V. Dimarogonas, ``Leveraging equivariances and symmetries in
  the control barrier function synthesis,'' 2025, arXiv:2509.04399.

\bibitem{zhang2018crown}
H.~Zhang, T.-W. Weng, P.-Y. Chen, C.-J. Hsieh, and L.~Daniel, ``Efficient
  neural network robustness certification with general activation functions,''
  in \emph{Advances in Neural Information Processing Systems (NeurIPS)}, 2018.

\bibitem{tayal2025cpncbf}
M.~Tayal, A.~Singh, P.~Jagtap, and S.~Kolathaya, ``{CP-NCBF}: A conformal
  prediction-based approach to synthesize verified neural control barrier
  functions,'' 2025, arXiv:2503.17395.

\bibitem{glotfelter2017nonsmooth}
P.~Glotfelter, J.~Cortés, and M.~Egerstedt, ``Nonsmooth barrier functions with
  applications to multi-robot systems,'' \emph{IEEE Control Systems Letters},
  vol.~1, no.~2, pp. 310--315, 2017.

\bibitem{breeden2023compositions}
J.~Breeden and D.~Panagou, ``Compositions of multiple control barrier functions
  under input constraints,'' in \emph{2023 American Control Conference (ACC)},
  2023, pp. 3688--3695.

\bibitem{grizzle1985symmetries}
J.~Grizzle and S.~Marcus, ``The structure of nonlinear control systems
  possessing symmetries,'' \emph{IEEE Transactions on Automatic Control},
  vol.~30, no.~3, pp. 248--258, 1985.

\bibitem{chen2021backup}
Y.~Chen, M.~Jankovic, M.~Santillo, and A.~D. Ames, ``Backup control barrier
  functions: Formulation and comparative study,'' in \emph{2021 60th IEEE
  Conference on Decision and Control (CDC)}, 2021, pp. 6835--6841.

\bibitem{so2024pncbf}
O.~So, Z.~Serlin, M.~Mann, J.~Gonzales, K.~Rutledge, N.~Roy, and C.~Fan, ``How
  to train your neural control barrier function: Learning safety filters for
  complex input-constrained systems,'' in \emph{2024 IEEE International
  Conference on Robotics and Automation (ICRA)}, 2024, pp. 11\,532--11\,539.

\bibitem{lee2010geometric}
T.~Lee, M.~Leok, and N.~H. McClamroch, ``Geometric tracking control of a
  quadrotor {UAV} on {SE(3)},'' in \emph{49th IEEE Conference on Decision and
  Control (CDC)}, 2010, pp. 5420--5425.

\bibitem{clopper1934}
C.~J. Clopper and E.~S. Pearson, ``The use of confidence or fiducial limits
  illustrated in the case of the binomial,'' \emph{Biometrika}, vol.~26, no.~4,
  pp. 404--413, 1934.

\bibitem{breeden2022sampled}
J.~Breeden, K.~Garg, and D.~Panagou, ``Control barrier functions in
  sampled-data systems,'' \emph{IEEE Control Systems Letters}, vol.~6, pp.
  367--372, 2022.

\end{thebibliography}
\end{document}